\pdfoutput=1  
\documentclass[11pt,a4paper,table,xcdraw]{article}
\usepackage[hyperref]{acl2021}
\usepackage{times}
\usepackage{latexsym}

\usepackage{microtype}

\aclfinalcopy 
\def\aclpaperid{34} 

\newenvironment{dense-itemize}{
\begin{itemize}
  \setlength{\itemsep}{1pt}
  \setlength{\parskip}{0pt}
  \setlength{\parsep}{0pt}
}{\end{itemize}}

\usepackage{graphicx}

\usepackage{booktabs}
\usepackage{multirow}

\usepackage{amsmath}
\usepackage{todonotes,enumitem}

\usepackage{soul}

\usepackage{afterpage}

\usepackage{gb4e}\noautomath

\newcommand{\kicktionary}{\textsc{Kicktionary}}

\usepackage[hyphenbreaks]{breakurl}

\title{\textsc{Kicktionary-LOME}: A Domain-Specific Multilingual \\ Frame Semantic Parsing Model for Football Language}

\author{Gosse Minnema\\
  Center for Language and Cognition \\
  University of Groningen, The Netherlands \\
  \texttt{g.f.minnema@rug.nl} 
    }
\date{}

\begin{document}
\maketitle

\begin{abstract}
    This technical report introduces an adapted version of the LOME frame semantic parsing model \cite{xia-etal-2021-lome} which is capable of automatically annotating texts according to the \kicktionary\ domain-specific framenet resource \cite{schmidt2009kicktionary}. Several methods for training a model even with limited available training data are proposed. While there are some challenges for evaluation related to the nature of the available annotations, preliminary results are very promising, with the best model reaching F1-scores of 0.83 (frame prediction) and 0.81 (semantic role prediction).
\end{abstract}
\section{Introduction}

Frame semantic parsing \cite{gildea-jurafsky, baker-etal-2007-semeval} is the task of automatically assigning frame semantic structures \cite{fillmore2006}, consisting of semantic frames and their associated semantic roles, to a text. Frame semantic parsers depend on the availability of language resources, called \textit{framenet}s, for providing the set of semantic frames and roles to be annotated, as well as a corpus of annotated examples to learn from.\footnote{Throughout this document, I will use \textit{framenet} (lower-case) for referring to any frame semantic language resource, and FrameNet (CamelCased) for the Berkeley FrameNet database.} The best-known framenet is Berkeley FrameNet \cite{baker2003}, a domain-general resource for English, but other framenets exist that are multilingual \cite{gilardi2018learning} or for specific other languages (e.g., FrameNet Brasil [\citealt{torrent2018towards}], ASFALDA French FrameNet [\citealt{candito-etal-2014-developing}], and many others) and for specific domains (e.g. \citealt{Venturi2009TowardsAF}, \citealt{datta-etal-2020-rad}).

This technical report will introduce a frame semantic parser for \kicktionary\ \cite{schmidt2009kicktionary}\footnote{See \url{http://kicktionary.de/}}, a domain-specific FrameNet covering the semantic domain of football (soccer), which provides semantic frame structures and lexical units (i.e., predicate words) with annotated examples in French, German, and English. As a basis for our system, we will use LOME \cite{xia-etal-2021-lome}, a recent multilingual frame semantic parsing model, as a basis, and experiment with several approaches for extending it to \kicktionary. 
\section{Methods}
\subsection{Dataset: \kicktionary} 
\kicktionary\ is organized around \textit{scenes} (abstract representations of the football semantic field), each of which contains a number of \textit{frames} (representation of a specific event, situation, or concept, associated with a set of semantic roles), each of which contains a number of \textit{lexical units} (\textit{LU}s; wordform-sense pairings functioning as predicates) in English, German, and French. 

\begin{table}[]
\centering
\resizebox{.48\textwidth}{!}{%
\begin{tabular}{@{}llllllllll@{}}
\toprule
\textbf{total} &  & \multicolumn{2}{l}{\textbf{by language}} &  & \multicolumn{2}{l}{\textbf{by frame (top-5)}} &  & \multicolumn{2}{l}{\textbf{by LU (top-5)}} \\ \midrule
8,342           &  & DE                 & 3,730                &  & Shot                      & 597               &  & treffen.v$^{\text{DE}}$                & 25              \\
               &  & EN                 & 2,374                &  & Pass                      & 492               &  & corner.n$^{\text{EN/FR}}$                 & 22              \\
               &  & FR                 & 2,239                &  & Goal                      & 448               &  & spielen.v$^{\text{DE}}$                & 20              \\
               &  &                    &                     &  & Player                    & 313               &  & passer.v$^{\text{FR}}$                 & 17              \\
               &  &                    &                     &  & Save                      & 256               &  & win.v$^{\text{EN}}$                    & 16              \\ \bottomrule
\end{tabular}
}
\caption{Counts of annotated examples in \kicktionary}\label{tab:numsents}
\end{table}

In turn, each lexical unit has a number of associated example sentences (\textit{exemplars}). Together, these form a corpus of 8,342 lexical units with semantic frame and role labels, annotated on top of 7,452 unique sentences (meaning that every sentence has, on average 1.11 annotated lexical units). Table~\ref{tab:numsents} gives more detailed information about the exemplars, split by relevant variables. 

\begin{table}[]
\resizebox{.48\textwidth}{!}{%
\begin{tabular}{@{}llll@{}}
\toprule
                          &                  & \textbf{Berkeley FN} & \textbf{Kicktionary} \\ \midrule
\multirow{2}{*}{fulltext} & unique sentences & 6,220                & 0                    \\
                          & annotated LUs    & 29,359               & 0                    \\ \midrule
\multirow{2}{*}{exemplar} & unique sentences & 163,801              & 7,452                \\
                          & annotated LUs    & 174,551              & 8,342                \\ \midrule
\multirow{2}{*}{total}    & unique sentences & 170,021              & 7,452                \\
                          & annotated LUs    & 203,910              & 8,342                \\ \bottomrule
\end{tabular}
}
\caption{Comparison of sentences and annotations in Berkeley FrameNet 1.7 and \kicktionary}\label{tab:annocounts}
\end{table}

For the purposes of using the annotations in \kicktionary\ for training a frame semantic parsing model, two properties of the corpus are likely to be important factors for the success of the training process: (i) the quantity of the data, but also (ii) the type of the corpus. In framenets, there are two possible types of corpora: \textit{fulltext} corpora, where entire documents are fully annotated (i.e., all possible predicates present in the text are annotated), and \textit{exemplar} corpora, which contain sentences that are specifically chosen to illustrate the semantics of particular predicates. Exemplar corpora typically have only one or a few annotations per sentence, which, depending on the exact task and the model that are used, can make learning from this data more challenging because there are many `gaps' (i.e., predicates that are included in the sentences but not annotated, which could cause problems especially when training end-to-end parsers); for this and other reasons, some developers of frame semantic parsers have chosen to ignore this kind of data altogether (e.g. \citealt{swayamdipta2017}). 
Table~\ref{tab:annocounts}\footnote{Examples are counted after preprocessing using the \texttt{bert-for-framenet} toolkit (\url{https://gitlab.com/gosseminnema/bert-for-framenet/}), it is possible that different ways of counting would yield slightly different numbers (e.g. due to sentence tokenization differences, discarding sentences with conflicting annotations, etc.)} compares both properties (i) and (ii) between the latest release of Berkeley FrameNet and \kicktionary. This comparison reveals two important limitations of \kicktionary: it contains only exemplar sentences, and much fewer of them than Berkeley FrameNet. On the other hand, given the limited semantic domain, and the limited number of frames in \kicktionary\ (Berkeley FrameNet has 1013 frames with at least one exemplar sentence versus only 106 in \kicktionary).

\subsection{Training \kicktionary-LOME}

LOME \cite{xia-etal-2021-lome} is a recent end-to-end\footnote{In this context, `end-to-end' means that the model performs all three traditional steps \cite{baker-etal-2007-semeval} of the frame semantic parsing process: target/predicate identification, frame identification, and semantic role identification. Until recently, there has not been much attention for frame semantic parsing as an end-to-end task; see \citet{minnema-nissim2021} for a recent study of training and evaluating semantic parsing models end-to-end.} frame semantic parsing model, and the only one (as far as I am aware) that can easily be used in a cross-lingual setting. While the authors of LOME do not provide a detailed evaluation of the model, they report a state-of-the art accuracy score on frame identification (on Berkeley FrameNet 1.7), a crucial component of the frame semantic parsing process.\footnote{At the time of publication; a model that was published later \cite{jiang-riloff-2021-exploiting} reported even stronger performance on the same metric, but is not capable of performing full, end-to-end frame semantic parsing and is not discussed any further here} 

LOME consists of a pre-trained XLM-R encoder \cite{conneau-etal-2020-unsupervised}, a BIO-tagger (for finding frame and role spans), and a typing module (for classifying frame and role labels). Thanks to the multilingual encoder, a trained LOME model can produce predictions for input texts in any of the 100 languages included in the XLM-R corpus, even if these languages are not present in the framenet training data.

To make the most out of the limited training data that is available from \kicktionary, three different training strategies were implemented:

\begin{itemize}
    \item \textbf{Simple:} follow standard LOME training procedure, training the decoders from scratch and fine-tuning the XLM-R encoder in the process;
    \item \textbf{Champions:} first fine-tune the XLM-R encoder on a masked language modeling task on a corpus\footnote{Collected by Albert Jan Schelhaas, a master student in the Groningen CL/NLP department, as part of his thesis project.} of 11,920 sentences of newspaper reports about UEFA Champions League matches. The motivation behind this approach is that this could help to `pre-adapt' the encoder towards the semantic domain of interest: if the language model has seen more examples of football language before it is exposed to the \kicktionary\ task, it might have learned better representations for this domained compared to the standard pre-trained XLM-R model.
    \item \textbf{Berkeley:} first train LOME on Berkeley FrameNet 1.7 following standard procedures; then, discard the decoder parameters but keep the fine-tuned XLM-R encoder. Finally, train the decoders on the \kicktionary\ dataset on top of this encoder. The intuition behind this technique is that encoder representations that are already `made useful' for doing frame semantic parsing (albeit on a different semantic domain) by fine-tuning on Berkeley FrameNet would make learning from the \kicktionary\ data more efficient. 
\end{itemize}

\subsection{Evaluation}

To be able to evaluate the model, the \kicktionary\ corpus was randomly split\footnote{Splitting was done on the unique sentence level to avoid having overlap in unique sentences between the training and evaluation sets. Splitting did not take into account frame and lexical unit labels; any given frame or LU can have zero or more instances in each of the splits.} into train (85\%), development (5\%), and test (10\%) sets. 

The fact that \kicktionary\ contains only exemplar (and no fulltext) sentences poses a challenge for evaluation: the trained LOME model will attempt to produce outputs for every possible predicate in the evaluation sentences, but since most sentences in the corpus have annotations for only one lexical unit per sentence, most of the outputs of the model cannot be evaluated: if the model produces a frame label for a predicate that was not annotated in the gold dataset, there is no way of knowing if a frame label should have been annotated for this lexical unit at all, and if so, what the correct label would have been. This implies that, given outputs from an (end-to-end) LOME system, one can reliably predict recall scores, but not precision scores. While it would be possible to partially circumvent this problem by providing LOME with the gold predicate(s) for each sentence and perform frame and role prediction for only these predicates, such a setup would be very unrealistic; it is very likely that in any real-world usage scenario of \kicktionary-LOME, the predicates to be annotated would not be known in advance. 

Instead, I will report two types of scores, which I hope will together give a rough estimate of the models' true precision:

\begin{itemize}
    \item \textbf{SCORE$_\text{RAW}$:} measure end-to-end accuracy given the model's predictions and the gold annotations. Because of the limited annotation coverage, we would expect very low precision scores, even for a hypothetical perfect model. However, these scores do say something about how `talkative' a model is in comparison to other models with similar recall: a lower precision score implies that the model predicts many `extra' labels beyond the gold annotations, while a higher score that fewer extra labels are predicted.
    \item \textbf{SCORE$_\text{GOLD\_PRED}$:} here, evaluation is restricted to only consider predicates that are present in the gold data. The precision score resulting from this will indicate how well a model does on predicting the correct frame and associated semantic roles, given a correctly identified predicate. This approach is similar to inputting gold predicates at inference time, but is more realistic because it only considers gold predicates that would be predicted by an end-to-end model.  
\end{itemize}

\section{Experiments}
\subsection{Implementation details}
LOME training was done using the same setting as in the original published model. Masked language model fine-tuning (for the Champions strategy) was done using the method described in \citet{bartl-etal-2020-unmasking}\footnote{I slightly adapted the system so that it works with XLM-R. See \url{https://github.com/marionbartl/gender-bias-BERT} for the original code.}, and was done over three epochs. LOME outputs confidence scores for each frame and role label that it assigns. Based on an initial manual inspection of dev-set predictions, I decided to apply a confidence threshold and only consider predictions with a confidence score of 1.0, in order to improve precision. 

All training was done on the University of Groningen's Peregrine cluster\footnote{\url{https://www.rug.nl/society-business/centre-for-information-technology/research/services/hpc/facilities/peregrine-hpc-cluster?lang=en}}, using a single NVIDIA V100 GPU. Training took between 3 and 8 hours per model, depending on the strategy. 

\subsection{Results and Discussion}
\begin{table*}[]
\centering
\resizebox{\textwidth}{!}{%
\begin{tabular}{@{}lcccccclccccccc@{}}
\toprule
 & \multicolumn{6}{c}{\textbf{frames}} &  & \multicolumn{6}{c}{\textbf{roles}} \\
\multicolumn{1}{c}{} & \multicolumn{3}{c}{\textbf{\textsc{raw}}} & \multicolumn{3}{c}{\textbf{\textsc{gold\_pred}}} & \textbf{} & \multicolumn{3}{c}{\textbf{\textsc{raw}}} & \multicolumn{3}{c}{\textbf{\textsc{gold\_pred}}} \\
\multicolumn{1}{c}{} & \textbf{R} & \textbf{P} & \textbf{F} & \textbf{R} & \textbf{P} & \textbf{F} &  & \textbf{R} & \textbf{P} & \textbf{F} & \textbf{R} & \textbf{P} & \textbf{F} \\ \midrule
Simple & \multicolumn{1}{r}{\cellcolor[HTML]{96FFFB}0.80} & \multicolumn{1}{r}{\cellcolor[HTML]{96FFFB}0.12} & \multicolumn{1}{r}{\cellcolor[HTML]{96FFFB}0.21} & \multicolumn{1}{r}{\cellcolor[HTML]{96FFFB}0.80} & \multicolumn{1}{r}{\cellcolor[HTML]{96FFFB}0.86} & \multicolumn{1}{r}{\cellcolor[HTML]{96FFFB}0.83} &  & \multicolumn{1}{r}{\cellcolor[HTML]{96FFFB}0.80} & \multicolumn{1}{r}{\cellcolor[HTML]{96FFFB}0.25} & \multicolumn{1}{r}{\cellcolor[HTML]{96FFFB}0.38} & \multicolumn{1}{r}{\cellcolor[HTML]{96FFFB}0.76} & \multicolumn{1}{r}{\cellcolor[HTML]{96FFFB}0.87} & \multicolumn{1}{r}{\cellcolor[HTML]{96FFFB}0.81} \\
Champions & \cellcolor[HTML]{C8E9D9}\textit{0.01} & \cellcolor[HTML]{FAE7E6}\textit{-0.03} & \cellcolor[HTML]{F8DBD9}\textit{-0.04} & \cellcolor[HTML]{C8E9D9}\textit{0.01} & \cellcolor[HTML]{E88981}\textit{-0.14} & \cellcolor[HTML]{F4C8C5}\textit{-0.06} &  & \cellcolor[HTML]{FAE8E7}\textit{-0.03} & \cellcolor[HTML]{F7D5D2}\textit{-0.05} & \cellcolor[HTML]{F5CAC7}\textit{-0.06} & \cellcolor[HTML]{FBEDEB}\textit{-0.02} & \cellcolor[HTML]{F8DDDB}\textit{-0.04} & \cellcolor[HTML]{FAE6E4}\textit{-0.03} \\
Berkeley & \cellcolor[HTML]{F7D5D3}\textit{-0.05} & \cellcolor[HTML]{9FD8BC}\textit{0.02} & \cellcolor[HTML]{7BCAA3}\textit{0.03} & \cellcolor[HTML]{F7D5D3}\textit{-0.05} & \cellcolor[HTML]{F6D4D1}\textit{-0.05} & \cellcolor[HTML]{F7D5D2}\textit{-0.05} &  & \cellcolor[HTML]{F4C8C5}\textit{-0.06} & \cellcolor[HTML]{57BB8A}\textit{0.04} & \cellcolor[HTML]{6BC398}\textit{0.04} & \cellcolor[HTML]{F7D5D2}\textit{-0.05} & \cellcolor[HTML]{7ECBA5}\textit{0.03} & \cellcolor[HTML]{FCF1F0}\textit{-0.02} \\ \bottomrule
\end{tabular}
}
\caption{Development set results. Colored cells indicate improvement (green) or worsening (red) with respect to the Simple model}\label{tab:results-dev}
\end{table*}

\begin{table*}[]
\centering
\resizebox{\textwidth}{!}{%
\begin{tabular}{@{}lcccccclccccccc@{}}
\toprule
 & \multicolumn{6}{c}{\textbf{frames}} &  & \multicolumn{6}{c}{\textbf{roles}} \\
\multicolumn{1}{c}{} & \multicolumn{3}{c}{\textbf{\textsc{raw}}} & \multicolumn{3}{c}{\textbf{\textsc{gold\_pred}}} & \textbf{} & \multicolumn{3}{c}{\textbf{\textsc{raw}}} & \multicolumn{3}{c}{\textbf{\textsc{gold\_pred}}} \\
\multicolumn{1}{c}{} & \textbf{R} & \textbf{P} & \textbf{F} & \textbf{R} & \textbf{P} & \textbf{F} &  & \textbf{R} & \textbf{P} & \textbf{F} & \textbf{R} & \textbf{P} & \textbf{F} \\ \midrule
Simple & \multicolumn{1}{r}{\cellcolor[HTML]{96FFFB}0.79} & \multicolumn{1}{r}{\cellcolor[HTML]{96FFFB}0.12} & \multicolumn{1}{r}{\cellcolor[HTML]{96FFFB}0.21} & \multicolumn{1}{r}{\cellcolor[HTML]{96FFFB}0.79} & \multicolumn{1}{r}{\cellcolor[HTML]{96FFFB}0.88} & \multicolumn{1}{r}{\cellcolor[HTML]{96FFFB}0.83} &  & \multicolumn{1}{r}{\cellcolor[HTML]{96FFFB}0.79} & \multicolumn{1}{r}{\cellcolor[HTML]{96FFFB}0.26} & \multicolumn{1}{r}{\cellcolor[HTML]{96FFFB}0.39} & \multicolumn{1}{r}{\cellcolor[HTML]{96FFFB}0.75} & \multicolumn{1}{r}{\cellcolor[HTML]{96FFFB}0.89} & \multicolumn{1}{r}{\cellcolor[HTML]{96FFFB}0.81} \\
Champions & \cellcolor[HTML]{A7DCC2}\textit{0.02} & \cellcolor[HTML]{FAE7E5}\textit{-0.03} & \cellcolor[HTML]{F8DAD8}\textit{-0.04} & \cellcolor[HTML]{A7DCC2}\textit{0.02} & \cellcolor[HTML]{E67C73}\textit{-0.15} & \cellcolor[HTML]{F4C6C2}\textit{-0.07} &  & \cellcolor[HTML]{F5FBF8}\textit{0.00} & \cellcolor[HTML]{F7D6D3}\textit{-0.05} & \cellcolor[HTML]{F5CECB}\textit{-0.06} & \cellcolor[HTML]{DCF1E6}\textit{0.01} & \cellcolor[HTML]{FBEBEA}\textit{-0.02} & \cellcolor[HTML]{FEFAFA}\textit{0.00} \\
Berkeley & \cellcolor[HTML]{FBECEB}\textit{-0.02} & \cellcolor[HTML]{97D5B7}\textit{0.03} & \cellcolor[HTML]{6CC499}\textit{0.04} & \cellcolor[HTML]{FBECEB}\textit{-0.02} & \cellcolor[HTML]{F8DEDB}\textit{-0.04} & \cellcolor[HTML]{FAE5E4}\textit{-0.03} &  & \cellcolor[HTML]{F6D4D1}\textit{-0.05} & \cellcolor[HTML]{67C295}\textit{0.04} & \cellcolor[HTML]{77C8A0}\textit{0.03} & \cellcolor[HTML]{F5CBC8}\textit{-0.06} & \cellcolor[HTML]{EEF9F4}\textit{0.00} & \cellcolor[HTML]{F9E1DF}\textit{-0.03} \\ \bottomrule
\end{tabular}
}
\caption{Test set results. Colored cells indicate improvement (green) or worsening (red) with respect to the Simple model}\label{tab:results-test}
\end{table*}

Results for the LOME models trained using the strategies specified in the previous sections are given in Table~\ref{tab:results-dev} (development set) and Table~\ref{tab:results-test} (test set). All scores were obtained using the \mbox{SeqLabel} evaluation method proposed in \citet{minnema-nissim2021}.

Given the limited availability of training data, the models' performance seems surprisingly good: recall scores on both frame and role prediction are close to 80\%, which is in fact considerably higher than published end-to-end performance (on the same metric and evaluation method) of previous frame semantic parsing models on Berkeley FrameNet (up to 70\% for frames, up to 40\% for roles; \citealt{minnema-nissim2021}).\footnote{No end-to-end scores on Berkeley FrameNet have been published yet for LOME; my own experiments indicate that LOME performs similarly to previous models on frame prediction, and improves on role prediction, with 56\% recall and 62\% precision on the test set.} As discussed above, precision is much harder to evaluate, but the results here still seem promising: $P_{\text{GOLD\_PRED}}$ is between 86-89\% on the Simple model. This means that, for predicates that the model \textit{should} annotate, it generally generates the correct frame and role predictions. What cannot be properly evaluated with currently available data is how frequently the model will `hallucinate', i.e., predict frame and role annotations for predicates that should not be annotated at all. 

The two strategies for making better use of (labeled and unlabeled) existing data resources, Champions and Berkeley, do not seem to yield much benefit. Champions gives a small improvement on frame recall, but this is offset by a large decrease in (gold predicate) precision. On the other hand, Berkeley improves on precision, but loses on recall. Thus, surprisingly, the Simple model, relying only on the limited training data available from \kicktionary, seems to be the best one overall. 

\section{Conclusions and Next Steps}

This technical report addressed the issue of training a frame semantic parser for \kicktionary, a domain-specific framenet resource for the domain of football language, and proposed strategies for adapting the LOME system \citet{xia-etal-2021-lome} for this purposes. While, given the available data, it is challenging to evaluate end-to-end performance, preliminary results are very promising. Future work on human evaluation on this data could be very helpful in getting a better estimate of the porposed models' true performance and usability in real-world applications. There is also still room for improvement of automatic evaluation; for example, work on checking to what degree LOME's predictions are consistent with the \kicktionary\ ontology.  Finally, adapting the SemEval'2007 method of evaluation \cite{baker-etal-2007-semeval} to the \kicktionary\ dataset could give a better indication of structure-level (as opposed to sequence label-level) performance.

\section*{Acknowledgements}
The research reported in this technical report was funded by
the Dutch National Science organisation (NWO) through
the project \textit{Framing situations in the Dutch language},
\texttt{VC.GW17.083/6215}. I would also like to thank Prof. Dr. Thomas Schmidt for giving me access to the original annotations from the \kicktionary\ project.

\bibliography{anthology,extra}
\bibliographystyle{acl_natbib}

\appendix

\end{document}